\documentclass[10pt,conference]{IEEEtran}
\usepackage{caption}
\usepackage{cite}

\usepackage{amsfonts}
\usepackage{hyperref}
\hypersetup{
    colorlinks=true,
    linkcolor=blue,
    filecolor=magenta,      
    urlcolor=blue,
    pdftitle={Overleaf Example},
    pdfpagemode=FullScreen,
    citecolor=blue
    }

\ifCLASSINFOpdf
   \usepackage[pdftex]{graphicx}
   \graphicspath{{figs/}}
   \DeclareGraphicsExtensions{.pdf,.jpeg,.png}
\else
   \usepackage[dvips]{graphicx}
   \graphicspath{{../figs/}}
   \DeclareGraphicsExtensions{.eps}
\fi
\usepackage[cmex10]{amsmath}

\usepackage{amsthm}

\usepackage{multicol}
\usepackage{algorithmic}
\usepackage{orcidlink}

\usepackage{array}

\ifCLASSOPTIONcompsoc
  \usepackage[caption=false,font=normalsize,labelfont=sf,textfont=sf]{subfig}
\else
  \usepackage[caption=false,font=footnotesize]{subfig}
\fi
\usepackage{float}

\usepackage{url}

\newif\iffinal
\finaltrue

\iffinal
\else
\usepackage[switch]{lineno}
\fi

\begin{document}

\title{Distributed Intelligent Video Surveillance for Early Armed Robbery Detection based on Deep Learning}

\iffinal

\author{
\IEEEauthorblockN{
Sergio Fernandez-Testa\IEEEauthorrefmark{1}\IEEEauthorrefmark{3}\orcidlink{0000-0003-3683-7497} 
and Edwin Salcedo\IEEEauthorrefmark{1}\IEEEauthorrefmark{3}\orcidlink{0000-0001-8970-8838}
}
\IEEEauthorblockA{\IEEEauthorrefmark{1} 
Centro de Investigación, Desarrollo e Innovación en Ingeniería Mecatrónica\\
Universidad Católica Boliviana ``San Pablo''\\
La Paz, Bolivia\\
Emails: sergiorodrigofernandeztesta@gmail.com, esalcedo@ucb.edu.bo}
\IEEEauthorblockA{\IEEEauthorrefmark{3} Both authors have contributed equally to this work.}
}


%

\else
\fi

\maketitle

\let\thefootnote\relax\footnote{\\979-8-3503-7603-6/24/\$31.00~\copyright~2024 IEEE\hfill}

\begin{abstract}
Low employment rates in Latin America have contributed to a substantial rise in crime, prompting the emergence of new criminal tactics. For instance, ``express robbery'' has become a common crime committed by armed thieves, in which they drive motorcycles and assault people in public in a matter of seconds. Recent research has approached the problem by embedding weapon detectors in surveillance cameras; however, these systems are prone to false positives if no counterpart confirms the event. In light of this, we present a distributed IoT system that integrates a computer vision pipeline and object detection capabilities into multiple end-devices, constantly monitoring for the presence of firearms and sharp weapons. Once a weapon is detected, the end-device sends a series of frames to a cloud server that implements a 3DCNN to classify the scene as either a robbery or a normal situation, thus minimizing false positives. The deep learning process to train and deploy weapon detection models uses a custom dataset with 16,799 images of firearms and sharp weapons. The best-performing model, YOLOv5s, optimized using TensorRT, achieved a final mAP of 0.87 running at 4.43 FPS. Additionally, the 3DCNN demonstrated 0.88 accuracy in detecting abnormal situations. Extensive experiments validate that the proposed system significantly reduces false positives while autonomously monitoring multiple locations in real-time. 
\end{abstract}

\IEEEpeerreviewmaketitle

\section{Introduction}
\label{sec:introduction}
Is it possible to detect an armed robbery in terms of seconds using closed-circuit television (CCTV)? Initially, one may think that this task is fairly trivial since pattern recognition in modern CCTV systems has reached high speeds, precision, and portability rates due to the advent of edge computing and low-cost embedded systems. On the other hand, one may realize that such a system would require a counterpart to confirm if a suspected scene is an authentic robbery. Nowadays, Latin American countries face significantly high crime rates, with armed robberies posing a prevalent threat to public safety. While CCTV-based security systems are becoming more common in Latin America, their primary function is to investigate crimes after they have happened, leaving victims vulnerable. Therefore, there is a constant need for active technologies to detect crimes in real-time. 

\begin{figure}[H]
    \centering
    \includegraphics[width=8cm]{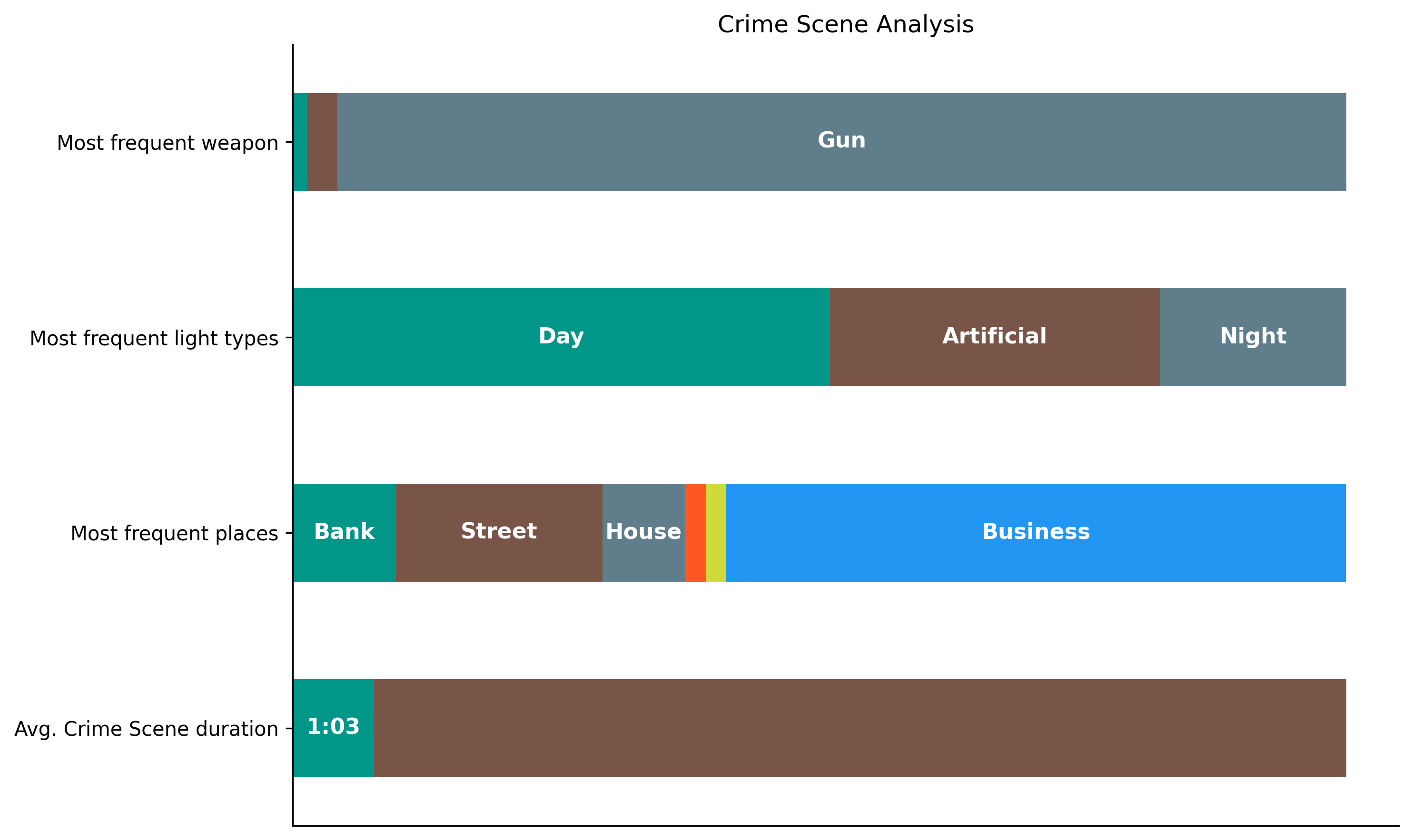}
    \caption{Summary of the analysis performed on 51 videos of robbery cases collected on social media.}
    \label{fig:vid-ana-dat}
\end{figure}

Armed robberies are generally carried out violently using firearms or sharp-edged weapons to intimidate victims and make them fear for their safety. To characterize this problem, we initially collected 51 videos from social networks (YouTube and Facebook), digital newspapers, and public repositories, resulting in a total duration of 2 hours, 40 minutes, and 24 seconds. The results, shown in Figure \ref{fig:vid-ana-dat}, illustrate important insights. For example, the average crime scene duration is 1 minute and 3 seconds, and guns are the most frequent weapon. Moreover, we also found that robberies tend to happen in business establishments more frequently by perpetrators who aim to unlawfully obtain something of value, such as money, jewels, or personal belongings. 

Deep learning, a subset of artificial intelligence (AI), has revolutionized various domains through its ability to analyze and interpret complex data patterns. Its application in crime detection, particularly armed robbery, offers a novel approach to enhancing public security. The advent of embedded devices and AI has enabled researchers to create innovative applications to detect firearms and sharp weapons in real time \cite{dugyala2023,ahmed2022,qi2021}. Moreover, some related investigations based on video surveillance, also known as active-based CCTV systems, extend the common recording function to actively detect different types of crime (e.g., arson, violence, robbery, among others) \cite{sultani2018,park2024} to alert the authorities promptly. 

Considering the problem and new technologies described above, this paper proposes an intelligent security system that uses computer vision and deep learning to detect firearms and sharp weapons on the edge, confirm genuine robberies on the cloud, and alert end-users. Moreover, the IoT (Internet of Things) system consists of an end device connected to the Internet, a web application for registering notifications and emergency users, and an Android app. While we used public datasets to create a new dataset to train the Yolo-based weapon detector, the data to train a robbery detector based on a 3DCNN came from multiple public sources, including social media and news media, available on the Internet. To deliver these contributions, we will proceed as follows. Section \ref{sec:relatedworks} briefly reviews the progress done on violence and weapon detection based on different sensing modalities, focusing on computer vision-based systems. Section \ref{sec:methods} describes the software and hardware components of the project, emphasizing the work done to distribute intelligence for video surveillance. Then, in Section \ref{sec:results}, we present and analyze the results. Finally, Section \ref{sec:conclusions} concludes the paper.  

\section{Related Works}
\label{sec:relatedworks}
 
Ongoing research into armed robbery detection systems is closely tied to weapon and intruder detection tasks, which are usually investigated using three types of sensing modalities: IoT-based, sound-based, and vision-based. The first group relies on sensors of different types to detect intruders, such as motion sensors (e.g. PIR,  microwave, and ultrasonic sensors), contact sensors installed in doors and windows, and pressure sensors placed on the floor. Despite the progress, these sensors have limitations related to the area they can monitor \cite{al2023acoustic}. Additionally, electromagnetic field sensing (EM) stands out because of its capability to detect concealed weapons by capturing the electromagnetic disruption caused by metal bodies. Authors in \cite{al2013emi} have explored combining machine learning and EM sensing to recognize weapons automatically; however, they highlighted further research is still needed to accurately differentiate weapons from cellphones, keys, or other metallic personal items, and minimize false positives.

Sound-based detection systems have been extensively explored to recognize shootings and intruders' speech in empty places using deep learning-based models \cite{khan2023}. For instance, Al-Khalli et al. \cite{al2023acoustic} proposed an Arduino device that incorporated a microphone module and an STA/LTA algorithm to monitor sudden noises or changes in audio, thus providing timely notifications to homeowners about potential intruders. Furthermore, researchers in \cite{wu2020} proposed a dataset with 4,754 untrimmed videos and audio signals corresponding to scenarios with abuses, explosions, fighting, riots, and shootings. The authors classified the dataset using a bi-modal neural network that simultaneously processed visual and audio data, achieving an accuracy of 78.64\%.

Vision-based systems for robbery detection are often associated with or incorporate violence detection techniques \cite{sultani2018, ruiz2021, velasco2021}. Approaches in this field commonly utilize RGB cameras \cite{wu2020, bhatti2021} or thermal cameras \cite{khor2024}, and tackle the pattern recognition challenge by analyzing individual images or sequential visual data. Furthermore, many studies utilize the vast video footage available from CCTV systems \cite{wu2020, bhatti2021}. Recent works have paved the way for automated armed robbery recognition by implementing gun, weapon, and knife detectors using deep learning models, such as YOLO, SSD, and EfficientDet \cite{dugyala2023, qi2021, ahmed2022}. Also, these models have been embedded inside stand-alone devices, to monitor people entering and exiting closed areas \cite{rojas2022}. For example, researchers in \cite{bhatti2021} gathered 15,313 trimmed CCTV videos of robberies and implemented YOLOv4 to detect firearms and sharp weapons.  

With the advent of high-performance computing (HPC), numerous studies focused on the processing and classification of extensive video datasets to detect violence behavior \cite{sultani2018, park2024}. On the one hand, this has been approached using end-to-end models, such as RNN \cite{traore2020}, 3DCNN \cite{park2024}, and LSTM \cite{abdali2019} models. For instance, Sultani et al. \cite{sultani2018} proposed a dataset with 1,900 long and untrimmed real-world surveillance videos. Then, the authors implemented CNN classifiers to discriminate video samples across 13 anomalies such as fighting, road accidents, burglary, robbery, vandalism, and arson, in a weakly-supervised manner. On the other hand, feature extractors (e.g. Mediapipe, OpenPose, Optical Flow \cite{park2024}) have been applied to obtain the key points of bodies and extract motion patterns present in frame sequences. Later, such feature representations were used to find anomalies in video feeds. Moreover, some studies have integrated both key point analysis and weapon detection to achieve improved results in determining whether a scene depicts a genuine robbery case \cite{ruiz2021, velasco2021}.   

Various aforementioned studies have explored the embedding of models inside end-devices to perform inference close to where data is collected \cite{al2023acoustic, rojas2022, khan2023, ahmed2022}. An example of this is presented by Qi et al. \cite{qi2021}, where a  HiSilicon-Hi3516 chip was used to detect, track, and classify moving objects with lightweight algorithms. Once an object was classified three times or more as a gun, the system sends the tracked object to the cloud in order to better classify it with a ResNet model. To the best of the author's knowledge, the work proposed in \cite{qi2021} is one of the few studies to explore inference distribution strategies to confirm twice the presence of guns. In contrast, the present proposal further explores this distribution approach by implementing modern object detectors, by reducing false positives by thresholding the cumulative detections' confidence, and by analyzing frame sequences to determine if there is an actual armed robbery being carried out in the scene.

\section{Proposed System}
\label{sec:methods}

\begin{figure}[H]
\centering
\includegraphics[trim=0cm 0cm 0cm 2cm, clip, width=0.58\textwidth]{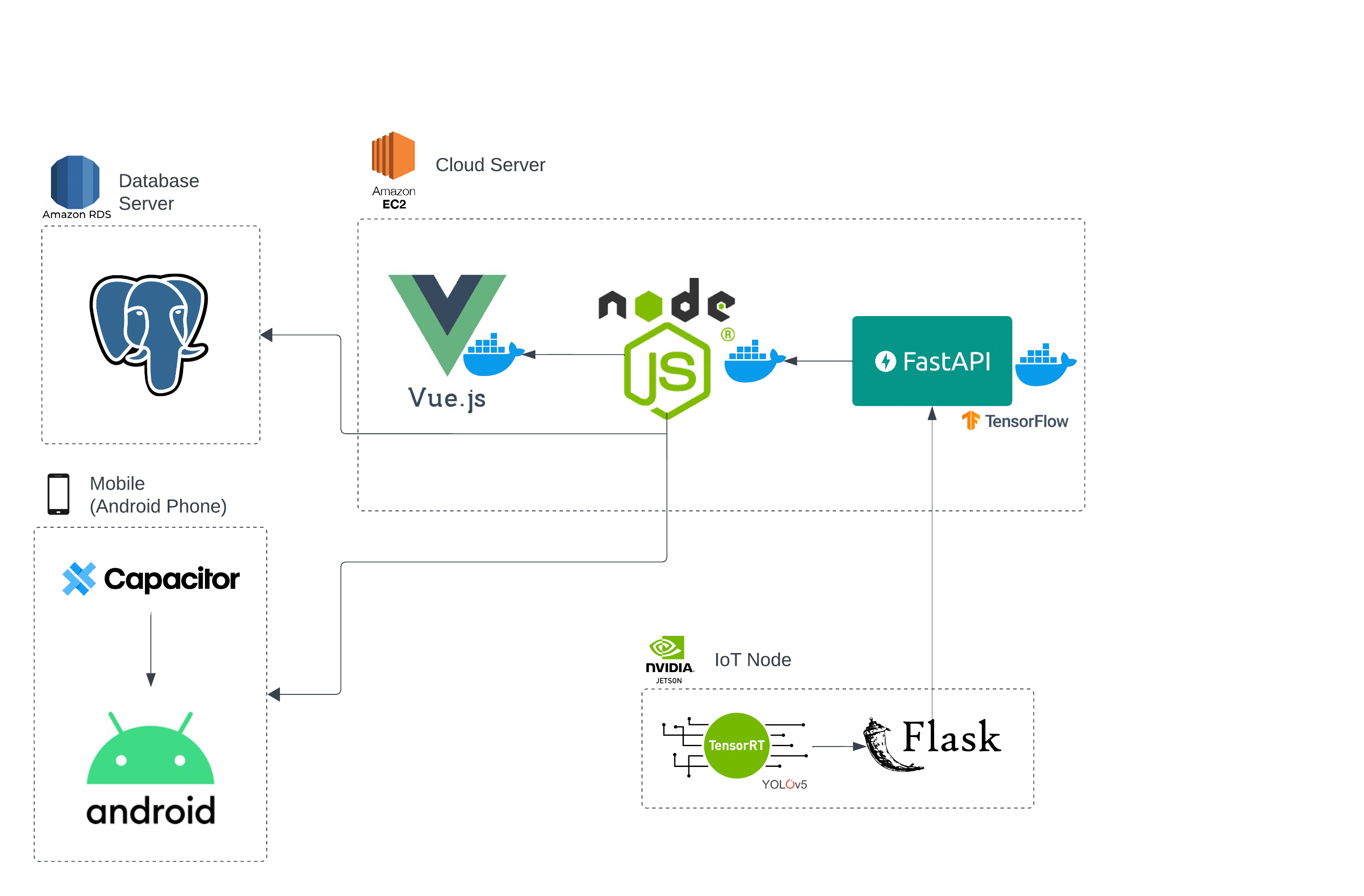}
\caption{General software architecture of the project.}
\label{fig:general-scheme}
\end{figure}

The project's goal was to develop an armed robbery detection system that could continuously monitor multiple locations and send notifications when a robbery has been confirmed. We achieve this by deploying weapon detectors into stand-alone video surveillance devices. Due to the detection challenges in real scenarios, we reduced false detections with a post-processing algorithm named \textit{Momentum}, which averages the weapon detections and obtains a final score about the presence of weapons. The momentum algorithm creates a confidence queue that includes the last 6 confidences of each class. After adding one element to the queue, the values are summed. The result is compared with a threshold, and if the sum is higher, the detection is considered valid. Moreover, we developed and deployed a 3DCNN model on a cloud server to receive a series of frames suspected as a robbery and confirm the happening of the event. Then, if a scene is confirmed as a robbery, notifications are sent via WhatsApp to the designated contacts. Figure \ref{fig:general-scheme} shows the general scheme of the project, which is composed of four main parts: a cloud server, a database, a mobile application, and end devices. These elements are explained in the following sections.

\subsection{Weapon Detection on the Edge}

\subsubsection{Data Preparation}
To create a new dataset for fire and white weapon detection, we used two object detection datasets available in the literature: SOHAS Weapon (SW) \cite{perez2020} and COCO Common Objects in Context (COCO) \cite{lin2014}. Both datasets contributed samples containing firearms and sharp weapons held by people or placed on a flat surface, but we obtained further samples from the COCO dataset to also incorporate background samples containing forks, remote controls, mobile phones, books, and toothbrushes, which are similar-looking objects to weapons. Once combined, the existing annotations were converted from VOC format to YOLO format. The resulting dataset had 9,720 images and included samples with annotated guns and knives. We continued the data processing stage applying data augmentation according to four techniques: cropping, rotating, generating mosaics, and varying the brightness. The resulting dataset contains approximately 16,799 images, which were later divided into three parts: training (70\%), validation (20\%), and testing (10\%). A more detailed summary of the object instances and different datasets is shown in Table \ref{tab:dataset}.

\begin{table}[H]
    \centering
    \caption{Number of instances per category for the initial combined dataset and the augmented one. \label{tab:dataset}}
    \begin{tabular}{|c|c|c|c|c|}
        \hline
        \textbf{Dataset/Class} & \textbf{Gun} & \textbf{Knife}  & \textbf{Background} & \textbf{Total} \\ \hline
        \text{SW} & 3,611 & 3,468 & 0 & 7,079 \\ \hline
        \text{COCO} & 0 & 0 & 2,641 & 2,641 \\ \hline
        \text{\textbf{Combined}} & 3,611 & 3,468 & 2,641 & \textbf{9,720} \\ \hline
        \text{Augmented} & 3,611 & 3,468 & 0 & 7,079 \\ \hline
        \text{\textbf{Total}} & 7,222 & 6,936 & 2,641 & \textbf{16,799} \\ \hline
    \end{tabular}
\end{table}

\subsubsection{Model Selection and Training} 

Due to their amenability for deployment in embedded platforms, different versions of YOLO (You Only Look Once) were implemented: YOLOv5, YOLOv7, YOLOS, and YOLOv8. Moreover, we trained the models using Google Colaboratory with a GPU Tesla A100. The models were evaluated using the metrics: mean Average Precision (mAP), area under the curve (AUC), accuracy (Acc), precision (P), recall (R), and F1-score (F1s). 
 







\subsubsection{Model Deployment}

The installation of the proposed end-devices requires a preliminary calibration step to adjust the brightness and contrast of each captured frame. This step is based on Equation \ref{eq:b_c_formula}, where $\alpha$  and $\beta$ modify the brightness and contrast, respectively.

\begin{equation}
    g(x,y) = \alpha \cdot f(x,y) + \beta
    \label{eq:b_c_formula}
\end{equation}

Next, the inference pipeline includes a post-processing algorithm, summarized as follows. Initially, a background removal step discriminates between static and moving regions by applying the Gaussian Mixture-based Background/Foreground Segmentation Algorithm \cite{kaewtrakulpong2002} implemented in OpenCV as \textit{BackgroundSubtractorMOG}. Then, inspired by the \textit{momentum} concept formulated in \cite{salcedo2024}, we implemented Equation \ref{eq:momentum} to add up the confidence of the detected weapons in a weighted manner. In this equation, $k$ is a parameter defined by the user in the unit interval, $c$ denotes a category processed at a time (i.e. $c \in \{gun, knife\}$), $t$ stands for time, and $q_{c}$ is the confidence level of each frame for a weapon category. Given a series of $n$ frames, the formula generates a \textit{momentum} score per category, which is compared to a threshold (1.05 for guns and 0.7 for knives) to determine if a sequence of frames should be sent to the Armed Robbery Detection model on the Cloud. Notably, by experimentally defining $n=5$ and $k=0.5$, we found that this approach reduced false positives in most deployments.

\begin{equation}
    momentum_{c} = \sum_{i=0}^{n} {q_{c}(t-i)} \cdot {k^{i}}
    \label{eq:momentum}
\end{equation}

\subsection{Armed Robbery Detection on the Cloud}

\subsubsection{Data Preparation}

This side of the detection system relies on a 3DCNN to verify the preliminary detections of weapons made on the edge. The dataset used to train the 3DCNN model was collected by combining videos from two sources: the UCF-Crime dataset \cite{sultani2018} and social media. Regarding the first source, we extracted the ``shooting'', ``robbery'', and ``normal'' categories from the dataset, making sure they contained real-world videos with visible weapons. Given that the UCF-Crime dataset consists of long videos with short periods showing the exact anomaly, we labeled manually the start and end seconds of the anomaly event. On the other hand, we downloaded public videos from Facebook, Twitter, YouTube, and other social networks. The search was performed using the following terms in their Spanish translation: ``assault'', ``armed'', and ``robbery''. 

The required preprocessing steps were defined by our findings with Exploratory Data Analysis (EDA) and by what we would require to correctly fit the data into the 3DCNN model. Thus, we initially normalized the videos' dimensions to 224 $\times$ 224 and converted them to grayscale using Python and OpenCV. We also made sure the videos had fewer than 600 frames (20 seconds) to avoid long samples, which in turn were split into several shorter videos to fit this duration. This resulted in 309 videos for both categories: ``robbery'' and ``normal''. Later, we applied data augmentation techniques to extend the dataset and avoid overfitting during model training. Specifically, we applied jointly two of the \textit{HorizontalFlip}, \textit{ShiftScaleRotate}, \textit{RandomBrightnessContrast}, \textit{CLAHE}, \textit{Affine}, and \textit{RandomResizedCrop} functions of the Albumentations library \cite{albumentations2020} to create new video samples. The augmented dataset contained 1,000 samples per category. 

To further evaluate the entire system, we collected an additional dataset named \textit{Simulation} dataset, which is available in \cite{simulationdataset}. This dataset consists of 40 videos of people armed with replicates of guns and knives, simulating armed robberies in outdoor and indoor scenarios. 

\subsubsection{3DCNN Architecture and Training}

\begin{figure}[H]
\centering
\includegraphics[width=8.5cm]{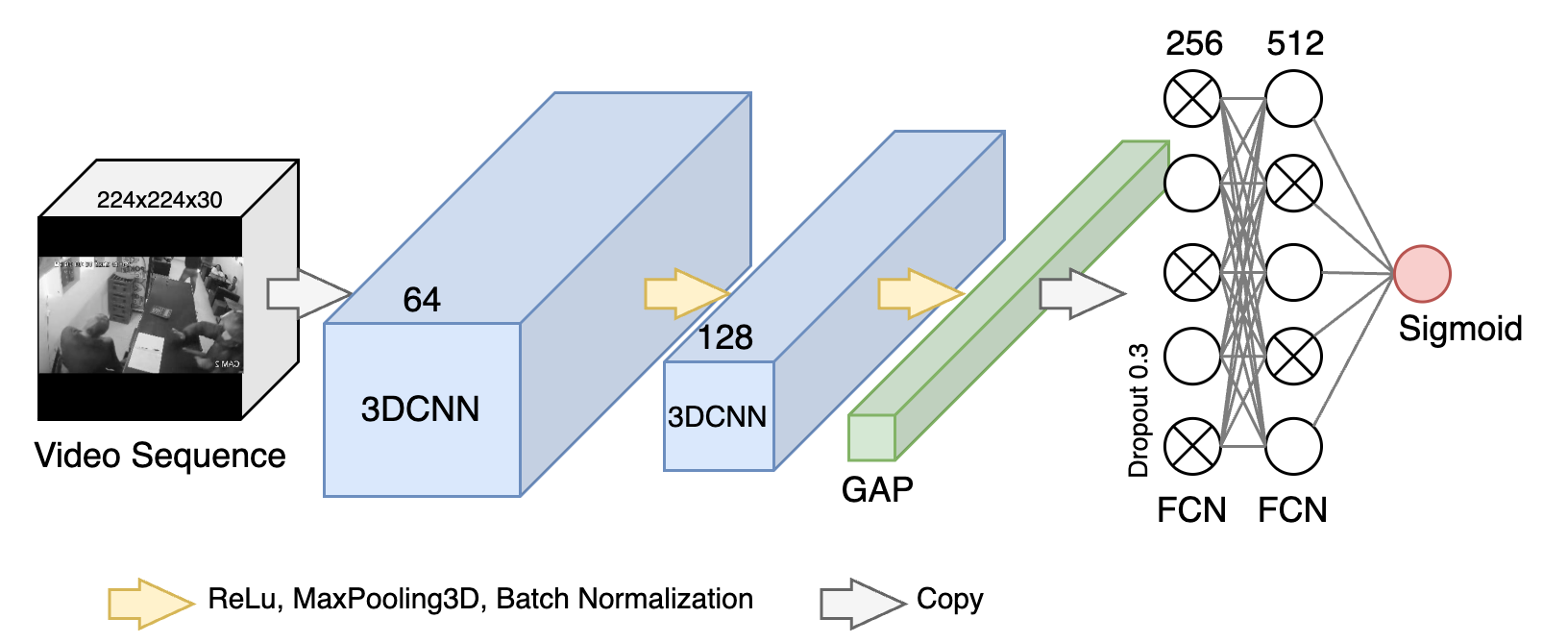}
\caption{Three-dimensional convolutional neural network (3DCNN).}
\label{fig:3dcnn}
\end{figure}

After multiple experiments, we obtained the final architecture (shown in Figure\ref{fig:3dcnn}) which consisted of two three-dimensional convolutional layers, a \textit{GlobalAveragePooling3D} layer, and a Fully Connected Network with two layers. We applied a Rectified Linear Unit (ReLU) as activation function in all hidden layers to avoid negative values. In the final layer, we combined Binary Cross Entropy (BCE) and Sigmoid functions to obtain a binary category. As for BCE, it was implemented according to Equation \ref{eq:bce}, where \textit{$y_i$} and \textit{$\hat{y_i}$} represents a ground truth binary classification vector and a predicted binary classification vector, respectively. Moreover, {\em T } is the number of data points, and the Sigmoid function {\em f} is shown in Equation \ref{eq:sigmoid}. Subsequently, we tested the system using Accuracy, F1-Score, Precision, and Recall. 

\begin{equation} 
\label{eq:bce}
BCE = -\frac{1}{T} \sum_{i=0}^{T} y_{i}\cdot log(f(\hat{y_{i}})) + (1 - y_{i})\cdot log(1 - f(\hat{y_{i}}))
\end{equation}

\begin{equation} 
\label{eq:sigmoid}
f(s_{i}) = \frac{1}{1+e^{-s_{i}}}
\end{equation}

\vspace{0.3cm}

\subsection{IoT Development}

\begin{figure}[H]
\centering
\includegraphics[trim=1cm 0cm 1cm 0cm, clip, width=8.5cm]{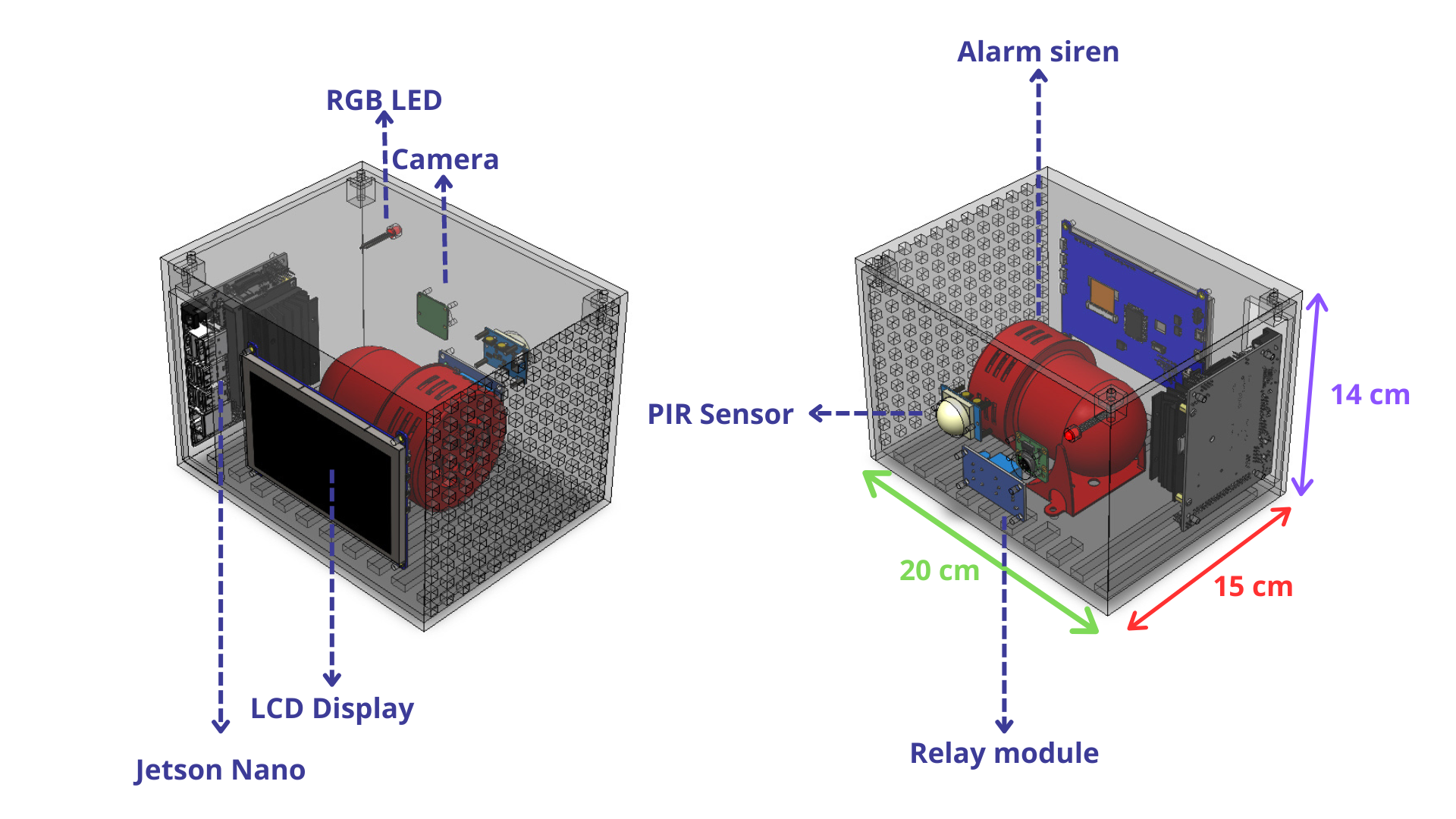}
\caption{Final design of the end device rendered with  SolidWorks.}
\label{fig:node_iot_render}
\end{figure}

\begin{figure}[H]
\centering
\includegraphics[trim=1cm 0cm 1cm 0cm, clip, width=7cm]{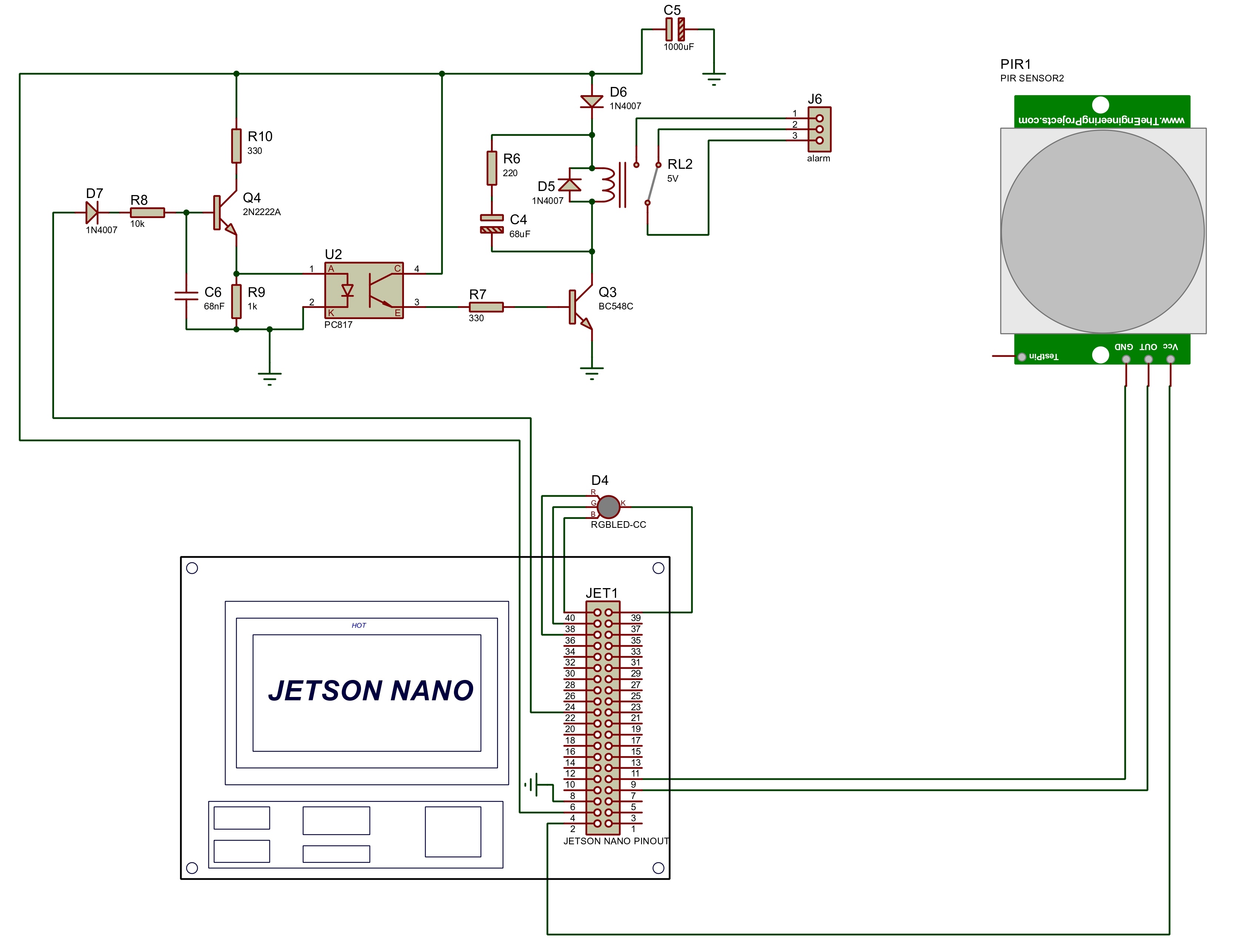}
\caption{Schematics of the end device.}
\label{fig:end-device-circuit}
\end{figure}

For IoT development, the web application's frontend was built using Vue.js, the backend using Express.js, and the database using PostgreSQL. The mobile application was not built from scratch, but rather using Capacitor.js, a framework that allows embedding web applications into mobile applications. Additionally, we implemented an end device consisting of a Jetson Nano running Flask and Tensor RT. When the device makes a detection, it saves the last 30 frames in GIF format and sends the file to the web server. The following is a detailed description of how the end device works. The end-device was prototyped using a 3D printed case, a Jetson Nano, a camera, an alarm siren, an LCD, a relay module, an RGB LED, and a Wi-Fi/Ethernet connection. These components are shown in a final 3D design of the end device in Figure \ref{fig:node_iot_render}. It is worth noting that the alarm siren is connected to a relay module to be activated via the Jetson Nano GPIO pins when a weapon is detected. This and other connections are further specified in Figure \ref{fig:end-device-circuit}. 

\vspace{-0.5cm}

\section{Experimental Results}
\label{sec:results}

\subsection{Weapon Detection}

\begin{figure}[H]
    \centering
    \includegraphics[width=0.48\textwidth]{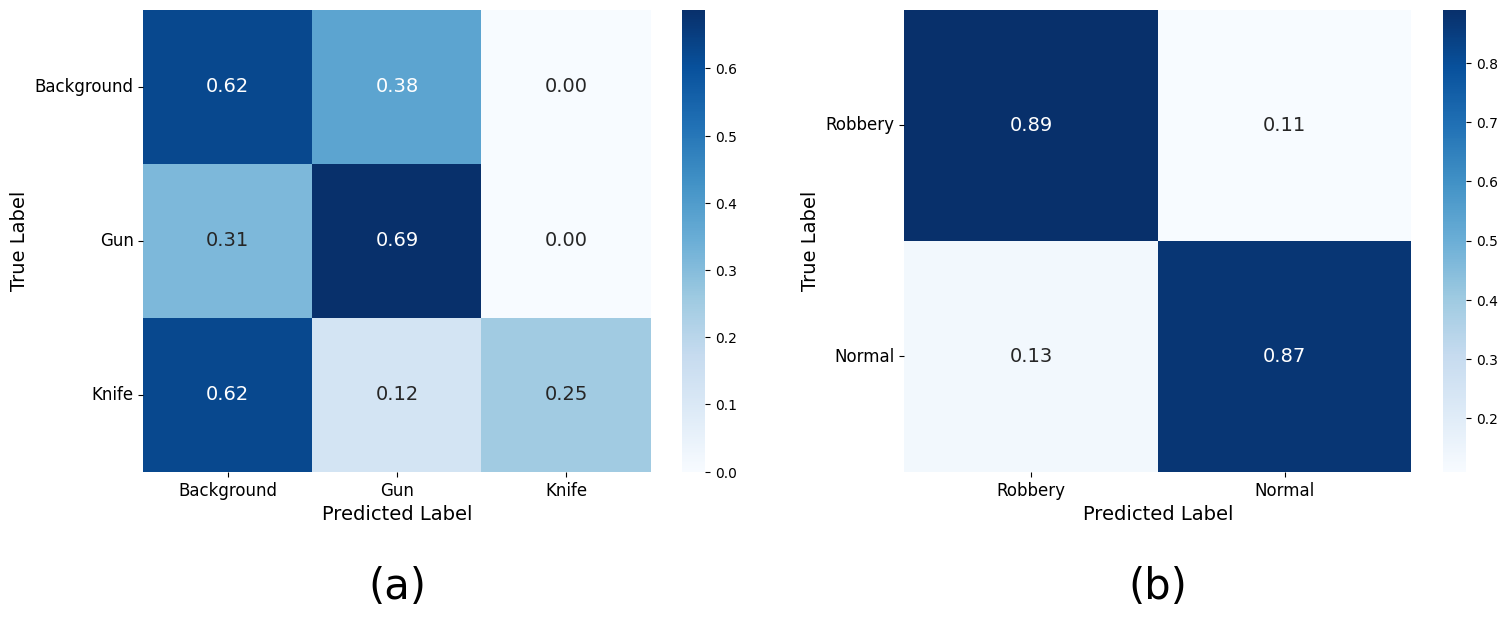}
    \caption{Confusion matrices of the YOLOv5s and 3DCNN models.}
    \label{fig:confusion_matrix}
\end{figure}

\begin{figure}[H]
    \centering
    \includegraphics[width=8.5cm]{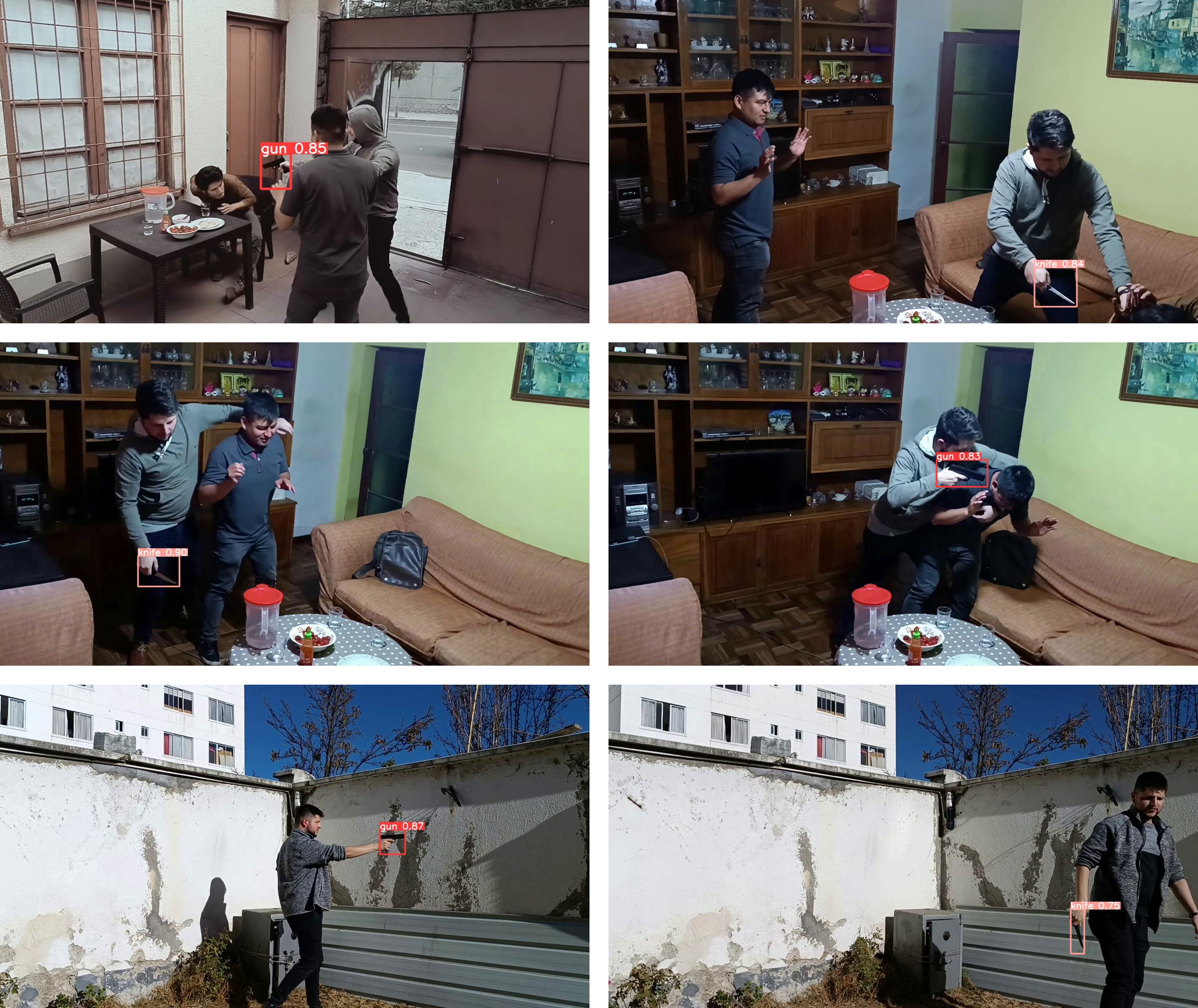}
    \caption{Inference results obtained using the end device.}
    \label{fig:samples}
\end{figure}

We processed the images, normalizing them to 640×480 dimensions and converting them to grayscale. Then, we split the final dataset into three subsets: training (70\%), validation (20\%), and testing (10\%).  After training and testing the YOLOv5, YOLOv7, YOLOS, and YOLOv8 models, we obtained the results shown in Table \ref{tab:metrics}. Notably, all models yielded very similar precision metrics; therefore, we also based the selection on the speed and deployment evaluations performed in \cite{terven2023} to finally select YOLOv5s. A confusion matrix illustrating the performance of this model and a set of detection samples are shown in Figures \ref{fig:confusion_matrix}(a) and \ref{fig:samples}, respectively.

Later, YOLOv5s was evaluated using the Simulation dataset \cite{simulationdataset} to test the impact of some pre-processing and post-processing techniques on the final inference. As shown in Table \ref{tab:res_exps}, the best results were obtained calibrating the brightness and contrast (B\&C), applying background removal (BR), compressing the model with Tensor RT (TRT), and implementing \textit{Momentum} (M). In the final row of this table, we include the results obtained when converting the frames to black and white (B\&W); however, the results were not optimal. It is worth noting that the results shown in Table \ref{tab:res_exps} were obtained considering only guns because we identified higher error rates when including robberies with knives. This is because knives are prone to disappearing due to light reflections and varying light sources, which highlights a challenge that can be addressed in future research. 

\subsection{Armed Robbery Detection}

The best-performing 3DCNN model achieved an overall accuracy of 0.88, demonstrating its ability to distinguish between robbery and non-robbery scenarios. Furthermore, its precision and recall values were 0.86 and 0.88, respectively. The obtained F1-score of 0.87 further emphasizes the model's robust performance in both detecting true robbery events and avoiding misclassifying normal activities. To further analyze the model's performance, we generated the confusion matrix shown in Figure \ref{fig:confusion_matrix}(b), which reveals that the model exhibits a higher rate of false negatives (0.13) compared to false positives (0.11). This suggests that the model is more prone to predicting normal situations, rather than detecting false alarms.

\begin{table}[H] 
\begin{center}
\caption{Experimental results obtained using state-of-the-art object detectors and the final dataset detailed in Table \ref{tab:dataset}.
\label{tab:metrics}}
\begin{tabular}{|c|c|c|c|c|}

\hline
\textbf{Metric} & YOLOv5 & YOLOv7 & YOLOS & YOLOv8 \\
\hline
\textbf{mAP} & 0.87 & 0.88 & 0.85 & 0.89 \\ 
    \textbf{Precision} & 0.84 & 0.85 & 0.82 & 0.86 \\ 
    \textbf{Recall} & 0.85 & 0.83 & 0.87 & 0.88 \\ 
    \textbf{F1-Score} & 0.84 & 0.84 & 0.85 & 0.87 \\ 
    \textbf{AUC} & 0.92 & 0.92 & 0.86 & 0.93 \\ 
\hline
\end{tabular}
\end{center}
\end{table}

\begin{table}[H]
\begin{center}
\caption{Performance results obtained by combining additional steps in the inference pipeline implemented in the end-device. \label{tab:res_exps}}
\begin{tabular}{|c|l|l|l|l|l|}

\hline
\# & Processes & Accuracy & F1-score & AUC  & FPS  \\
\hline
1      & B\&C                 & 0.48  & 0.41  & 0.56 & 2.21 \\ 
2      & B\&C, BR             & 0.5   & 0.49  & 0.57 & 2.16 \\
3      & B\&C, BR, TRT        & 0.75  & 0.74  & 0.74 & 4.43 \\
4      & B\&C, BR, TRT, B\&W  & 0.62  & 0.61  & 0.61 & 4.37 \\ 
\hline

\end{tabular}
\end{center}
\end{table}


\begin{figure}[H]
    \centering
    \includegraphics[trim=0cm 0cm 0cm 0cm, clip, width=0.5\textwidth]{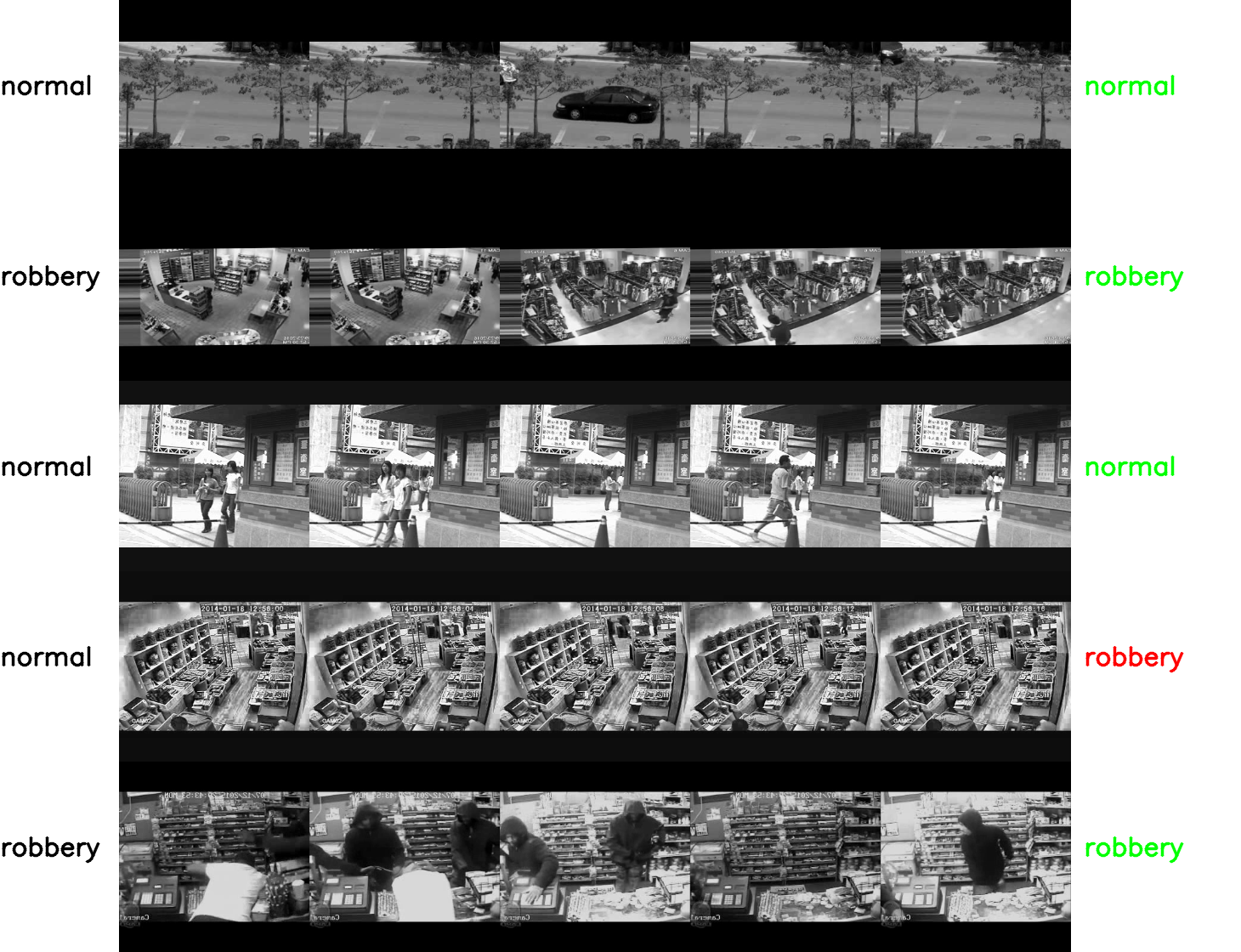}
    \caption{Results obtained by the 3DCNN. Real labels are shown on the left and predictions on the right.}
    \label{fig:samples3DCNN}
\end{figure}

To check misclassifications, we programmed a script to randomly select a number of video samples defined by the user, use the best-performing 3DCNN, and visualize the videos' frames alongside the ground truth and predicted labels. This method allowed us to observe either no misclassifications or, at most, one misclassification per five video samples analyzed in each script execution. A sample of these visualizations is shown in Figure \ref{fig:samples3DCNN}, from which we can notice that the model misclassified the fourth sample (fourth row) because it associated a single object moving suddenly in the scene with a robbery. This highlights an area for future improvement and testing. Yet, most importantly, we evaluated the 3DCNN with frames sent by an end device to test the entire system. Table \ref{tab:res_3dcnn} shows the results obtained by the 3DCNN model, which, in the best case, achieved an accuracy of 0.88, sending 15 frames per second during two seconds to the cloud.

\begin{table}[H]
\begin{center}
\caption{Performance results obtained by the 3DCNN model when processing frames sent by the algorithm deployed inside the end device. The second and third columns indicate the frame frequency considered for inference with the 3DCNN. \label{tab:res_3dcnn}}
\begin{tabular}{|c|c|c|c|c|c|}
\hline
\# & FPS & Seconds & Accuracy & F1-score & AUC  \\
\hline
1 & 15 & 2 & 0.88 & 0.87 & 0.88 \\
2 & 6  & 5 & 0.82 & 0.53 & 0.68 \\
3 & 5  & 6 & 0.82 & 0.58 & 0.71 \\
4 & 1  & 30 & 0.6  & 0.2  & 0.47 \\
5 & 0.5 & 60 & 0.57 & 0.19 & 0.45 \\
6 & 0.25 & 120 & 0.67 & 0.38 & 0.58 \\
\hline
\end{tabular}
\end{center}
\end{table}

\section{Conclusions}
\label{sec:conclusions}

In conclusion, this project has succeeded in developing a distributed armed robbery detection system based on computer vision and edge computing. On the edge, YOLOv5s was selected as the most suitable after evaluating several YOLO models. Additionally, YOLOv5s, implemented on a Jetson Nano card using Tensor RT, yielded promising results when the camera input was further processed by varying the brightness and contrast, removing the background, and adding the confidence of weapon detections using the Momentum algorithm. It is worth noting that knife detection has room for improvement, since knives are affected by varying lighting conditions and reflections.

On the other hand, the inclusion of a robbery detection model in the cloud allowed us to reduce false positives. This was based on a 3DCNN and obtained 0.88 accuracy in detecting robbery scenes consisting of 30 frames. However, the 3DCNN's relatively high false-negative rate and misclassifications suggest a potential area for improvement. The complete approach allowed us to incorporate a notification system that works via WhatsApp and sends alerts about genuine robberies that are detected in around 2 seconds. Future work could focus on exploring techniques to leverage video generation tools to increase the training dataset. Additionally, incorporating extra contextual information, such as audio cues or object interactions, could further enhance the system's accuracy.

\bibliographystyle{IEEEtran}

\bibliography{main}

\begin{thebibliography}{10}
\providecommand{\url}[1]{#1}
\csname url@samestyle\endcsname
\providecommand{\newblock}{\relax}
\providecommand{\bibinfo}[2]{#2}
\providecommand{\BIBentrySTDinterwordspacing}{\spaceskip=0pt\relax}
\providecommand{\BIBentryALTinterwordstretchfactor}{4}
\providecommand{\BIBentryALTinterwordspacing}{\spaceskip=\fontdimen2\font plus
\BIBentryALTinterwordstretchfactor\fontdimen3\font minus \fontdimen4\font\relax}
\providecommand{\BIBforeignlanguage}[2]{{%
\expandafter\ifx\csname l@#1\endcsname\relax
\typeout{** WARNING: IEEEtran.bst: No hyphenation pattern has been}%
\typeout{** loaded for the language `#1'. Using the pattern for}%
\typeout{** the default language instead.}%
\else
\language=\csname l@#1\endcsname
\fi
#2}}
\providecommand{\BIBdecl}{\relax}
\BIBdecl

\bibitem{dugyala2023}
R.~Dugyala, M.~V.~V. Reddy, C.~T. Reddy, and G.~Vijendar, ``Weapon detection in surveillance videos using yolov8 and pelsf-dcnn,'' in \emph{E3S Web of Conferences}, vol. 391.\hskip 1em plus 0.5em minus 0.4em\relax EDP Sciences, 2023, p. 01071.

\bibitem{ahmed2022}
S.~Ahmed, M.~T. Bhatti, M.~G. Khan, B.~L{\"o}vstr{\"o}m, and M.~Shahid, ``Development and optimization of deep learning models for weapon detection in surveillance videos,'' \emph{Applied Sciences}, vol.~12, no.~12, p. 5772, 2022.

\bibitem{qi2021}
D.~Qi, W.~Tan, Z.~Liu, Q.~Yao, and J.~Liu, ``A dataset and system for real-time gun detection in surveillance video using deep learning,'' pp. 667--672, 2021.

\bibitem{sultani2018}
W.~Sultani, C.~Chen, and M.~Shah, ``Real-world anomaly detection in surveillance videos,'' \emph{2018 IEEE/CVF Conference on Computer Vision and Pattern Recognition}, 2018.

\bibitem{park2024}
J.-H. Park, M.~Mahmoud, and H.-S. Kang, ``Conv3d-based video violence detection network using optical flow and rgb data,'' \emph{Sensors}, vol.~24, no.~2, p. 317, 2024.

\bibitem{al2023acoustic}
N.~Al-Khalli, S.~Alateeq, M.~Almansour, Y.~Alhassoun, A.~B. Ibrahim, and S.~A. Alshebeili, ``Real-time detection of intruders using an acoustic sensor and internet-of-things computing,'' \emph{Sensors}, vol.~23, no.~13, p. 5792, 2023.

\bibitem{al2013emi}
A.~Al-qubaa, ``An electromagnetic imaging system for metallic object detection and classification,'' Ph.D. dissertation, Newcastle University, 2013.

\bibitem{khan2023}
T.~Khan, ``Towards an indoor gunshot detection and notification system using deep learning,'' \emph{Applied System Innovation}, vol.~6, no.~5, p.~94, 2023.

\bibitem{wu2020}
P.~Wu, J.~Liu, Y.~Shi, Y.~Sun, F.~Shao, Z.~Wu, and Z.~Yang, ``Not only look, but also listen: Learning multimodal violence detection under weak supervision,'' in \emph{Computer Vision--ECCV 2020: 16th European Conference, Glasgow, UK, August 23--28, 2020, Proceedings, Part XXX 16}.\hskip 1em plus 0.5em minus 0.4em\relax Springer, 2020, pp. 322--339.

\bibitem{ruiz2021}
J.~Ruiz-Santaquiteria, A.~Velasco-Mata, N.~Vallez, G.~Bueno, J.~A. Alvarez-Garcia, and O.~Deniz, ``Handgun detection using combined human pose and weapon appearance,'' \emph{IEEE Access}, vol.~9, pp. 123\,815--123\,826, 2021.

\bibitem{velasco2021}
A.~Velasco-Mata, J.~Ruiz-Santaquiteria, N.~Vallez, and O.~Deniz, ``Using human pose information for handgun detection,'' \emph{Neural Computing and Applications}, vol.~33, no.~24, pp. 17\,273--17\,286, 2021.

\bibitem{bhatti2021}
M.~T. Bhatti, M.~G. Khan, M.~Aslam, and M.~J. Fiaz, ``Weapon detection in real-time cctv videos using deep learning,'' \emph{IEEE Access}, vol.~9, pp. 34\,366--34\,382, 2021.

\bibitem{khor2024}
W.~Khor, Y.~K. Chen, M.~Roberts, and F.~Ciampa, ``Infrared thermography as a non-invasive scanner for concealed weapon detection,'' in \emph{Defence Secur. Doctoral Symp. 2023}, 2024.

\bibitem{rojas2022}
W.~Rojas, E.~Salcedo, and G.~Sahonero, ``Adras: airborne disease risk assessment system for closed environments,'' in \emph{Annual International Conference on Information Management and Big Data}.\hskip 1em plus 0.5em minus 0.4em\relax Springer, 2022, pp. 96--112.

\bibitem{traore2020}
A.~Traoré and M.~A. Akhloufi, ``Violence detection in videos using deep recurrent and convolutional neural networks,'' in \emph{2020 IEEE International Conference on Systems, Man, and Cybernetics (SMC)}, 2020, pp. 154--159.

\bibitem{abdali2019}
A.-M.~R. Abdali and R.~F. Al-Tuma, ``Robust real-time violence detection in video using cnn and lstm,'' in \emph{2019 2nd Scientific Conference of Computer Sciences (SCCS)}, 2019, pp. 104--108.

\bibitem{perez2020}
F.~P{\'e}rez-Hern{\'a}ndez, S.~Tabik, A.~Lamas, R.~Olmos, H.~Fujita, and F.~Herrera, ``Object detection binary classifiers methodology based on deep learning to identify small objects handled similarly: Application in video surveillance,'' \emph{Knowledge-Based Systems}, vol. 194, p. 105590, 2020.

\bibitem{lin2014}
T.-Y. Lin, M.~Maire, S.~Belongie, J.~Hays, P.~Perona, D.~Ramanan, P.~Doll{\'a}r, and C.~L. Zitnick, ``Microsoft coco: Common objects in context,'' in \emph{Computer Vision--ECCV 2014: 13th European Conference, Zurich, Switzerland, September 6-12, 2014, Proceedings, Part V 13}.\hskip 1em plus 0.5em minus 0.4em\relax Springer, 2014, pp. 740--755.

\bibitem{kaewtrakulpong2002}
P.~KaewTraKulPong and R.~Bowden, ``An improved adaptive background mixture model for real-time tracking with shadow detection,'' \emph{Video-based surveillance systems: Computer vision and distributed processing}, pp. 135--144, 2002.

\bibitem{salcedo2024}
E.~Salcedo, Y.~Uchani, M.~Mamani, and M.~Fernandez, ``Towards continuous floating invasive plant removal using unmanned surface vehicles and computer vision,'' \emph{IEEE Access}, 2024.

\bibitem{albumentations2020}
A.~Buslaev, V.~I. Iglovikov, E.~Khvedchenya, A.~Parinov, M.~Druzhinin, and A.~A. Kalinin, ``Albumentations: Fast and flexible image augmentations,'' \emph{Information}, vol.~11, no.~2, 2020.

\bibitem{simulationdataset}
\BIBentryALTinterwordspacing
Authors, ``Simulation dataset,'' Gdrive, accessed: Aug. 16, 2024. [Online]. Available: \url{https://drive.google.com/drive/folders/18_gmn4qohngT7pbf0JBTlW7DemHEBowp}
\BIBentrySTDinterwordspacing

\bibitem{terven2023}
J.~Terven, D.-M. C{\'o}rdova-Esparza, and J.-A. Romero-Gonz{\'a}lez, ``A comprehensive review of yolo architectures in computer vision: From yolov1 to yolov8 and yolo-nas,'' \emph{Machine Learning and Knowledge Extraction}, vol.~5, no.~4, pp. 1680--1716, 2023.

\end{thebibliography}

\end{document}